\documentclass[10pt,twocolumn,letterpaper]{article}

\usepackage{cvpr}
\usepackage{times}
\usepackage{epsfig}
\usepackage{graphicx}
\usepackage{amsmath}
\usepackage{amssymb}

\usepackage[utf8]{inputenc} 
\usepackage[T1]{fontenc}    
\usepackage{booktabs}       
\usepackage{amsfonts}       
\usepackage{nicefrac}       
\usepackage{microtype}      
\usepackage{caption}

\usepackage[export]{adjustbox}
\usepackage{tabularx,ragged2e}
\usepackage{multirow}

\newcolumntype{C}{>{\Centering\arraybackslash}X}  
\newcolumntype{b}{>{\Centering}X}
\newcolumntype{s}{>{\Centering\hsize=.5\hsize}X}
\newcolumntype{j}{>{\Centering\hsize=.4\hsize}X}
\newcolumntype{k}{>{\Centering\hsize=.3\hsize}X}
\newcolumntype{y}{>{\Centering\hsize=.2\hsize}X}

\makeatletter
\def\blfootnote{\xdef\@thefnmark{}\@footnotetext}
\makeatother

\usepackage[breaklinks=true,letterpaper=true,colorlinks,bookmarks=false]{hyperref}

\cvprfinalcopy 
\def\cvprPaperID{6054} 

\begin{document}

\title{Dynamic Face Video Segmentation via Reinforcement Learning}

\author{
Yujiang Wang$^{1,2*}$ \hspace{0.4cm} 
Mingzhi Dong$^{3}$ \hspace{0.4cm} 
Jie Shen$^{1,2\dagger}$ \hspace{0.4cm}  
Yang Wu$^{4}$ \\
Shiyang Cheng$^{2}$ \hspace{0.4cm} 
Maja Pantic$^{1,2}$\\
$^{1}$Imperial College London \hspace{0.4cm} 
$^{2}$Samsung AI Center Cambridge \\
$^{3}$University College London \hspace{0.4cm} 
$^{4}$Nara Institue of Science and Technology \\
{\tt\small yujiang.wang14@imperial.ac.uk, mingzhi.dong.13@ucl.ac.uk, jie.shen07@imperial.ac.uk} \\
{\tt\small 
yangwu@rsc.naist.jp,
shiyang.c@samsung.com,
m.pantic@imperial.ac.uk} 
}

\maketitle


\begin{abstract}

\blfootnote{$^{\dagger}$Corresponding author.}
\blfootnote{$^{*}$Yujiang Wang conducted this research during his internship at Samsung AI Center, Cambridge and Nara Institute of Science and Technology}

For real-time semantic video segmentation, most recent works utilised a dynamic framework with a key scheduler to make online key/non-key decisions. Some works used a fixed key scheduling policy, while others proposed adaptive key scheduling methods based on heuristic strategies, both of which may lead to suboptimal global performance. To overcome this limitation, we model the online key decision process in dynamic video segmentation as a deep reinforcement learning problem and learn an efficient and effective scheduling policy from expert information about decision history and from the process of maximising global return. Moreover, we study the application of dynamic video segmentation on face videos, a field that has not been investigated before. By evaluating on the 300VW dataset, we show that the performance of our reinforcement key scheduler outperforms that of various baselines in terms of both effective key selections and running speed. Further results on the Cityscapes dataset demonstrate that our proposed method can also generalise to other scenarios. To the best of our knowledge, this is the first work to use reinforcement learning for online key-frame decision in dynamic video segmentation, and also the first work on its application on face videos.  
\end{abstract}

\section{Introduction}
\label{introduction}
In computer vision, semantic segmentation is a computationally intensive task which performs per-pixel classification on images. Following the pioneering work of Fully Convolutional Networks (FCN) \cite{long2015fully}, tremendous progress has been made in recent years with the propositions of various deep segmentation methods \cite{chen2018deeplab, badrinarayanan2017segnet, wu2019wider, zhao2017pyramid, autodeeplab2019, deeplabv3plus2018, lin2017refinenet, paszke2016enet,chen2017rethinking, zhao2018icnet, nekrasov2018light}. To achieve accurate result, these image segmentation models usually employ heavy-weight deep architectures and additional steps such as spatial pyramid pooling  \cite{zhao2017pyramid,chen2017rethinking,chen2018deeplab} and multi-scaled paths of inputs/features  \cite{chen2016attention,zhao2018icnet,lin2016efficient,chen2018deeplab, chen2014semantic, lin2017refinenet, pohlen2017full}, which further increase the computational workload. For real-time applications such as autonomous driving, video surveillance, and facial analysis \cite{wang2018face}, it is impractical to apply such methods on a per-frame basis, which will result in high latency intolerable to those applications. Therefore, acceleration becomes a necessity for these models to be applied in real-time video segmentation.  

\begin{figure}[ht!]
  \begin{center}
  \includegraphics[width=.96\linewidth]{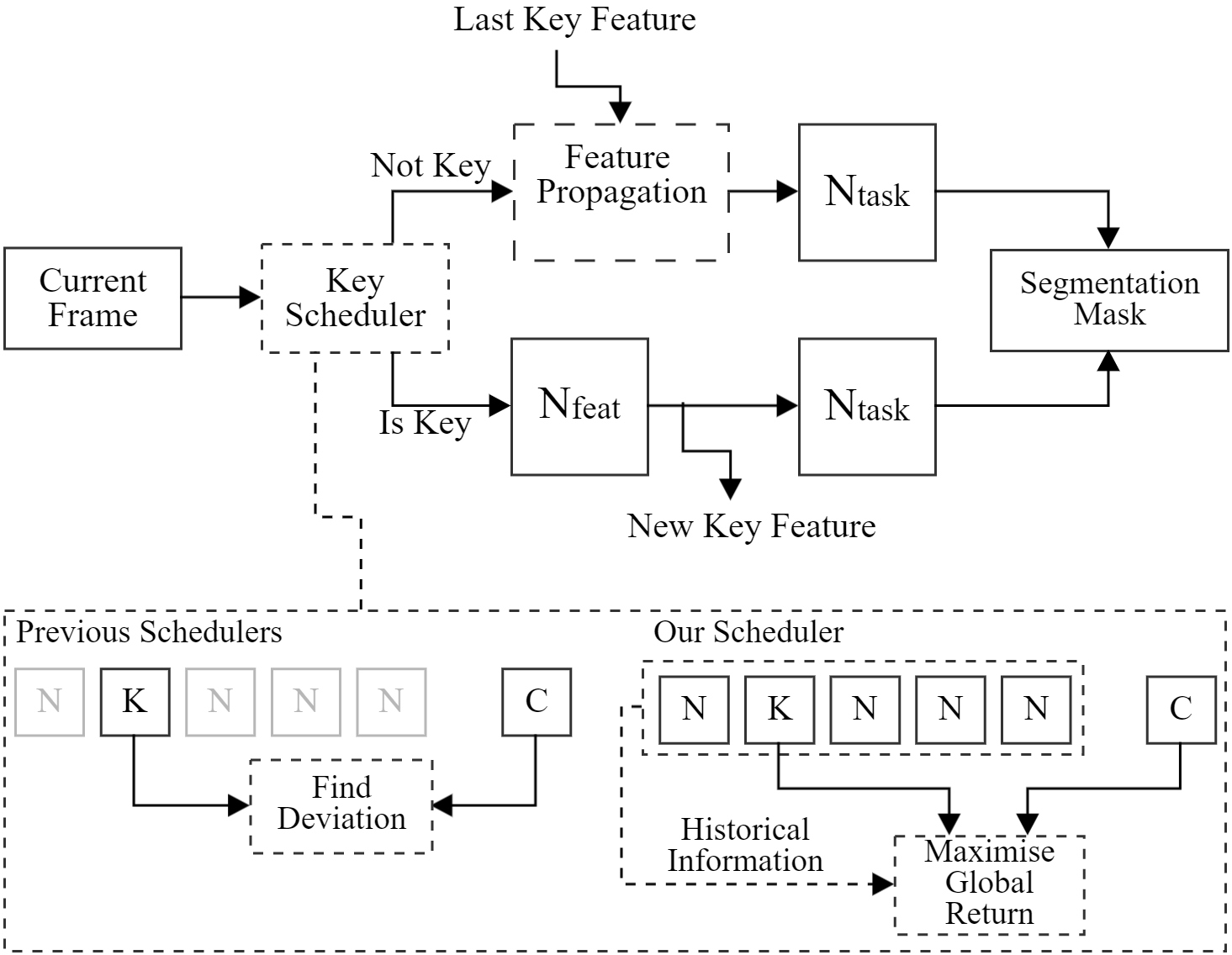}
  \end{center}
  \vspace{-2mm}
  \caption{Up: The dynamic video segmentation framework in which a key scheduler is used to make online key/non-key predictions. Bottom: a comparison between previous key schedulers and ours. Previous works only consider deviations between current frame (C) and the last key frame (K), while our scheduler takes into account C, K and historical information from non-key frames (N), aiming to maximise the global return. } 
  \label{fig:commonFramework}
  \vspace{-2mm}
\end{figure}

Various methods \cite{shelhamer2016clockwork, zhu2017deep, xu2018dynamic, li2018low, jain2018inter, nekrasov2019architecture, jain2018accel, nilsson2018semantic, gadde2017semantic} have been proposed to accelerate video segmentation. Because adjacent frames in a video often share a large proportion of similar pixels, most of these works utilise a dynamic framework which separates frames into key and non-key frames and produce their segmentation masks differently. As illustrated in Fig. \ref{fig:commonFramework} (up), a deep image segmentation model $\mathcal{N}$ is divided into a heavy feature extraction part $\mathcal{N}_{feat}$ and a light task-related part $\mathcal{N}_{task}$. To produce segmentation masks, key frames would go through both $\mathcal{N}_{feat}$ and $\mathcal{N}_{task}$, while a fast feature interpolation method is used to obtain features for the non-key frames by warping $\mathcal{N}_{feat}$'s output on the last key frame (LKF), thus to avoid the heavy cost of running $\mathcal{N}_{feat}$ on every frame. On top of that, a key scheduler is used to predict whether an incoming frame should be a key or non-key frame. 

As an essential part of dynamic video segmentation, decisions made by the key scheduler could significantly affect the overall performance \cite{li2018low,xu2018dynamic,zhu2018towards} of the video segmentation framework. However, this topic is somewhat underexplored by the community. Recent works have either adopted a fixed key scheduler \cite{nekrasov2019architecture,zhu2017deep,jain2018accel,jain2018inter}, or  proposed adaptive schedulers \cite{xu2018dynamic,li2018low,zhu2018towards} which are trained to heuristically predict similarities (or deviations) between two video frames. Those key schedulers lack awareness of the global video context and can lead to suboptimal performance in the long run.

To overcome this limitation, we propose to apply Reinforcement Learning (RL) techniques to expose the key scheduler to the global video context. Leveraging additional expert information about decision history, our scheduler is trained to learn key-decision policies that maximise the long-term returns in each episode, as shown in Fig. \ref{fig:commonFramework} (bottom).

We further study the problem of dynamic face video segmentation with our method. Comparing to semantic image/video segmentation, segmentation of face parts is a less investigated field \cite{gucclu2017end,zhou2015interlinked,kae2013augmenting,smith2013exemplar,nirkin2018face,warrell2009labelfaces,scheffler2011joint,yacoob2006detection,lee2008markov, ghiasi2015using}, and there are fewer works on face segmentation in videos \cite{wang2018face,saito2016real}. Existing works either used engineered features \cite{kae2013augmenting, smith2013exemplar,warrell2009labelfaces,scheffler2011joint,yacoob2006detection,lee2008markov}, or employed outdated image segmentation models like FCN \cite{long2015fully} on a per-frame basis \cite{nirkin2018face, saito2016real, wang2018face} without a dynamic acceleration mechanism. Therefore, we propose a novel real-time face segmentation system utilising our key scheduler trained by Reinforcement Learning (RL). 

We evaluate the performances of the proposed method on the 300 Videos in the Wild (300VW) dataset \cite{shen2015first} for the task of real-time face segmentation. Comparing to several baseline approaches, we show that our reinforcement key scheduler can make more effective key-frame decisions at the cost of fewer resource. Through further experiment conducted on the Cityscapes dataset \cite{cordts2016cityscapes} for the task of semantic urban scene understanding, we demonstrate that our method can also generalise to other scenarios.

\section{Related works}
\label{RelaedWorks}

\textbf{Semantic image segmentation} \hspace{0.1cm} Fully Convolutional Networks (FCN) \cite{long2015fully} is the first work to use fully convolutional layers and skip connections to obtain pixel-level predictions for image segmentation. Subsequent works have made various improvements, including the usage of dilated convolutions \cite{chen2014semantic,chen2018deeplab,chen2017rethinking,yu2015multi,yu2017dilated}, encoder-decoder architecture \cite{badrinarayanan2017segnet,lin2017refinenet,deeplabv3plus2018}, Conditional Random Fields (CRF) for post-processing \cite{zheng2015conditional,chen2014semantic,chen2018deeplab}, spatial pyramid pooling to capture multi-scale features \cite{zhao2017pyramid,chen2018deeplab,chen2017rethinking} and Neural Architecture Search (NAS) \cite{zoph2016neural} to search for the best-performing architectures \cite{dpc2018, autodeeplab2019}. Nonetheless, such models usually require intensive computational resources, and thus may lead to unacceptably high latency in video segmentation.



\textbf{Dynamic video segmentation} \hspace{0.1cm} Clockwork ConvNet \cite{shelhamer2016clockwork} promoted the idea of dynamic segmentation by fixing part of the network. Deep Feature Flow (DFF) \cite{zhu2017deep} accelerated video recognition by leveraging optical flow (extracted by FlowNet \cite{zhu2017flow, ilg2017flownet} or SpyNet \cite{ranjan2017optical}) to warp key-frame features. Similar ideas are explored in \cite{xu2018dynamic, jain2018accel, nilsson2018semantic, gadde2017semantic}. Inter-BMV~\cite{jain2018inter} used block motion vectors in compressed videos for acceleration. Mahasseni \etal \cite{mahasseni2017budget} employed convolutions with uniform filters for feature interpolation, while Li \etal \cite{li2018low} used spatially-variant convolutions instead. Potential interpolation architectures were searched in \cite{nekrasov2019architecture}. 

On the other hand, studies of key schedulers are comparatively rare. Most existing works adopted fixed key schedulers \cite{nekrasov2019architecture,zhu2017deep,jain2018accel,jain2018inter}, which is inefficient for real-time segmentation. Mahasseni \etal \cite{mahasseni2017budget} suggested a budget-aware, LSTM-based key selection strategy trained with reinforcement learning, which is only applicable for \textit{offline} scenarios. DVSNet \cite{xu2018dynamic} proposed an adaptive key decision network based on the similarity score between the interpolated mask and the key predictions, i.e., low similarity scores leading to new keys and vice versa. Similary, Li \etal \cite{li2018low} introduced a dynamic key scheduler trained to predict the deviations between two video frames by the inconsistent low-level features, and \cite{zhu2018towards} proposed to adaptively select key frames depending on the pixels with inconsistent temporal features. Those adaptive key schedulers only consider deviations between two frames, and therefore lack understandings of global video context, leading to suboptimal performances. 

\textbf{Semantic face segmentation} \hspace{0.1cm} Semantic face parts segmentation received far less attention than that of image/video segmentation. Early works on this topic mostly used engineered features \cite{kae2013augmenting, smith2013exemplar,warrell2009labelfaces,scheffler2011joint,yacoob2006detection,lee2008markov} and were designed for static images. Saito \etal \cite{saito2016real} employed graphic cut algorithm to refine the probabilistic maps from a FCN trained with augmented data. In \cite{nirkin2018face}, a semi-supervised data collection approach was proposed to generate more labelled facial images with random occlusions to train FCN. Recently, Wang \etal \cite{wang2018face} integrated Conv-LSTM \cite{xingjian2015convolutional} with FCN \cite{long2015fully} to extract face masks from video sequence, while the run-time speed did not improve. None of the se works considered to adopt video dynamics for accelerations, and we are the first to do so for real-time face segmentation. 

\textbf{Reinforcement learning} \hspace{0.1cm} In model-free Reinforcement Learning (RL), an agent receives a state $\mathbf{s}_t$ at each time step $t$ from the environment, and learns a policy $\pi_\theta(a_j|\mathbf{s}_t)$ with parameters $\theta$ that guides the agent to take an action $a_j \in \mathcal{A}$ to maximise the cumulative rewards $J=\sum_{t=1}^{\infty} \gamma^{t-1}r_t$. RL has demonstrated impressive performance on various fields such as robotics and complicated strategy games \cite{lillicrap2015continuous,silver2016mastering,mnih2013playing,vinyals2017starcraft,silva2017moba, vinyals2019alphastar}. In this paper, we show that RL can be seamlessly applied to online key decision problem in real-time video segmentation, and we chose the policy gradient with reinforcement \cite{williams1992simple} to learn $\pi_\theta$, where gradient ascend was used for maximising the objective function $J_\pi (\theta)$. 

\section{Methodology}
\label{Methodology}

\begin{figure*}[t]
  \begin{center}
  \includegraphics[width=.92\linewidth]{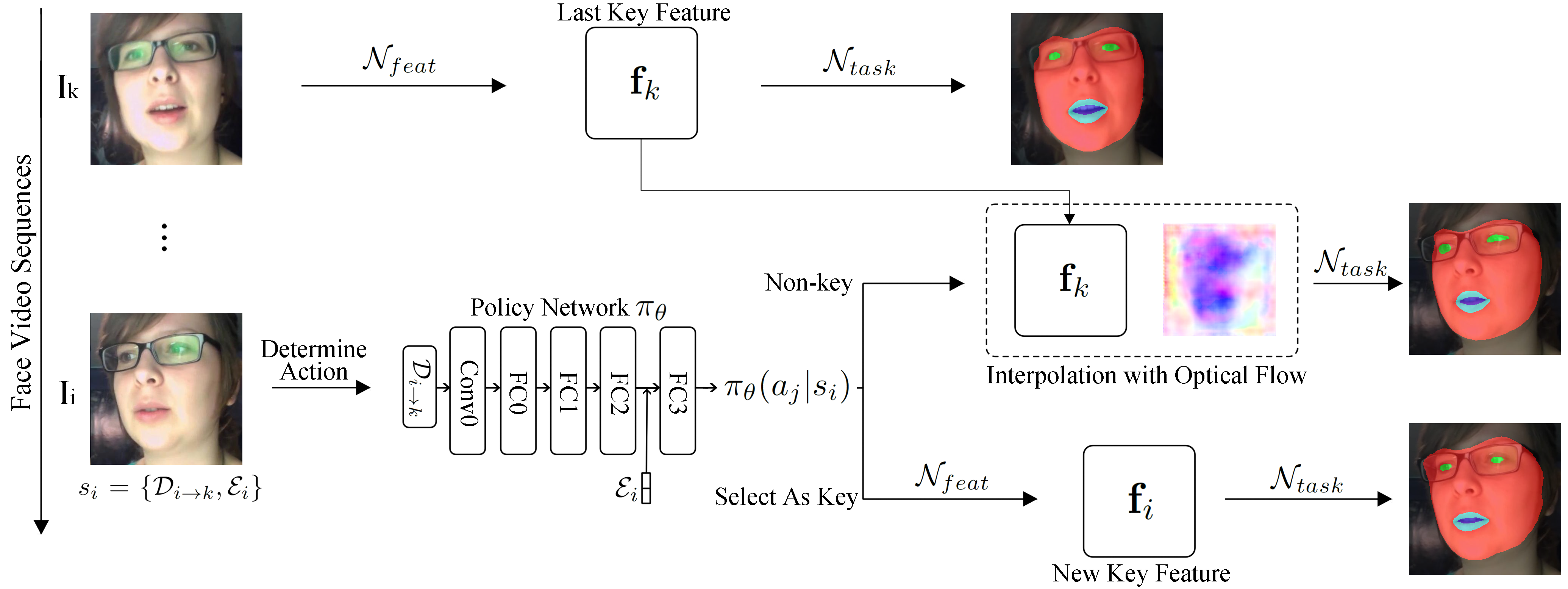}
  \end{center}
  \vspace{-2mm}
\caption{An overview of our system. $\mathbf{I}_k$ is the last key frame (key decision process not shown) with feature $\mathbf{f}_k$ extracted by $\mathcal{N}_{feat}$. For an incoming frame $\mathbf{I}_i$, its input state $\mathbf{s}_i$ includes two components: the deviation information $\mathcal{D}_{i \rightarrow k}$ between $\mathbf{I}_i$ and $\mathbf{I}_k$, and the expert information $\mathcal{E}_{i}$ about decision history. $\mathcal{D}_{i \rightarrow k}$ is fed into Conv0 layer of policy network $\pi_\theta$, while $\mathcal{E}_{i}$ is concatenated to the output of FC2 layer. Basing on $\mathbf{s}_i$, $\pi_\theta$ gives probabilities output $\pi_\theta(a_j|\mathbf{s}_i)$ regarding taking key or non-key actions. For a non-key action, the optical flow between $\mathbf{I}_i$ and $\mathbf{I}_k$ will be used to warp $\mathbf{f}_k$ to $\mathbf{f}_i$, while for a key action, $\mathbf{I}_i$ will go through $\mathcal{N}_{feat}$ to obtain a new key feature $\mathbf{f}_i$.}
  \label{fig:system_overview}
  \vspace{-2mm}
\end{figure*}

\subsection{System Overview} 
Our target is to develop an efficient and effective key scheduling policy $\pi_\theta(\mathbf{a}|\mathbf{s})$ for the dynamic video segmentation system. To this end, we use Deep Feature Flow \cite{zhu2017deep} as the feature propagation framework, in which the optical flow is calculated by a light-weight flow estimation model $\mathcal{F}$ such as FlowNet \cite{zhu2017flow,ilg2017flownet} or SpyNet \cite{ranjan2017optical}. Specifically, an image segmentation model $\mathcal{N}$ can be divided into a time-consuming feature extraction module $\mathcal{N}_{feat}$ and a task specified module $\mathcal{N}_{task}$. We denote the last key frame as $\mathbf{I}_k$ and its features extracted by $\mathcal{N}_{feat}$ as $\mathbf{f}_k$, i.e., $\mathbf{f}_k=\mathcal{N}_{feat}(\mathbf{I}_k)$.
For an incoming frame $\mathbf{I}_i$, if it is a key frame, the feature is $\mathbf{f}_i=\mathcal{N}_{feat}(\mathbf{I}_i)$ and the segmentation mask is $\mathbf{y}_i=\mathcal{N}_{task}(\mathbf{f}_i)$; if not, instead of using the resource-intensive module $\mathcal{N}_{feat}$ for feature extraction, its feature $\mathbf{f}_i$ will be propagated by a feature interpolation function $\mathcal{W}$, which involves the flow field $\mathbf{M}_{i \rightarrow k}$ from $\mathbf{I}_i$ to $\mathbf{I}_k$, the scale field $\mathbf{S}_{i \rightarrow k}$ from $\mathbf{I}_i$ to $\mathbf{I}_k$, and key frame feature $\mathbf{f}_k$, hence the predicted mask becomes $\mathbf{y}_i = \mathcal{N}_{task}(\mathbf{f}_i)$. Please check \cite{zhu2017deep} for more details on the feature propagation process.

On top of the DFF framework, we design a light-weight policy network $\pi_\theta$ to make online key predictions. The state $\mathbf{s}_i$ at frame $\mathbf{I}_i$ consists of two parts, the deviation information $\mathcal{D}_{i \rightarrow k}$ which describes the differences between $\mathbf{I}_k$ and $\mathbf{I}_i$, and the expert information $\mathcal{E}_{i}$ regarding key decision history (see Section \ref{ReinforcementTraining} for details), i.e., $\mathbf{s}_i=\{\mathcal{D}_{i \rightarrow k},\mathcal{E}_{i}\}$. Feeding $\mathbf{s}_i$ as input, the policy network outputs the action probabilities $\pi_\theta(a_j|\mathbf{s}_i)$ where $a_j \in \{a_0,a_1\}$ and $\pi_\theta(a_0|\mathbf{s}_i)+\pi_\theta(a_1|\mathbf{s}_i)=1.0$ (we define $a_0$ for non-key action and $a_1$ for the key one). For an incoming frame $\mathbf{I}_i$, if $\pi_\theta(a_1|\mathbf{s}_t)>\tau$ where $\tau$ is a threshold, it will be identified as a key frame, vice versa. In general, key action $a_1$ will lead to a segmentation mask with better quality than the ones given by action $a_0$.

In this work, we utilise the FlowNet2-s model \cite{ilg2017flownet} as the optical flow estimation function $\mathcal{F}$. DVSNet \cite{xu2018dynamic} has shown that the high-level features from FlowNet models contain sufficient information about the deviations between two frames, and it can also be easily fetched along with optical flow without additional cost. Therefore, we adopt the features of FlowNet2-s model for $\mathcal{D}_{i \rightarrow k}$. It is worthwhile to notice that by varying $\mathcal{D}_{i \rightarrow k}$ properly, our key scheduler can be easily integrated into other dynamic segmentation frameworks \cite{jain2018accel,li2018low,nekrasov2019architecture,jain2018inter,zhu2018towards} which do not use optical flow. Fig. \ref{fig:system_overview} gives an overview of our system.

\subsection{Training Policy Network} 
\label{ReinforcementTraining}

\textbf{Network structure} \hspace{0.1cm} Our policy network comprises of one convolution layer and four fully connected (FC) layers. The FlowNet2-s feature  $\mathcal{D}_{i \rightarrow k}$ is fed into the first convolution layer Conv0 with 96 channels, followed by FC layers (FC0, FC1 and FC2) with output size being 1024, 1024 and 128 respectively. Two additional channels containing expert information about decision history $\mathcal{E}_{i}$ are concatenated to the output of FC2 layer. The first channel records the Key All Ratio (KAR), which is the ratio between the key frame and every other frames in decision history, while the second channel contains the Last Key Distance (LKD), which is the interval between the current and the last key frame. KAR provides information on the frequency of historical key selection, and LKD gives awareness about the length of continuous non-key decisions. Hence, the insertion of KAR and LKD extends the output dimension of FC2 to 130, while FC3 layer summarises all these information and gives action probabilities $\pi_\theta(a_j|\mathbf{s}_i)$ where $a_j \in \{a_0,a_1\}$, $a_0$ and $a_1$ stand for non-key and key action correspondingly.

\textbf{Reward definition} \hspace{0.1cm} We use mean Intersection-over-Union (mIoU) as the metric to evaluate the segmentation masks. We denote the mIoU of $\mathbf{y}_i$ from a non-key action $a_0$ as $U^i_{a_0}$, the mIoU from key action $a_1$ as $U^i_{a_1}$, and the reward $r_i$ at frame $\mathbf{I}_i$ is defined in Eq. \ref{eq:1}. Such definition encourages the scheduler to choose key action on the frames that would result in larger improvement over non-key action, and it also reduces the variances of mIoUs across the video.


%
\begin{align}
\begin{split}
\label{eq:1}
\displaystyle
r_i &= 
    \begin{cases}
        0, & a_j=a_0.\\
      U^i_{a_1} - U^i_{a_0}, & a_j=a_1. \\
    \end{cases} \\
\end{split}
\end{align}

If no groundtruth is available (such that mIoU could not be calculated), we use the segmentation mask from key action as the pseudo groundtruth mask. In this case, the reward formulation is changed to Eq. \ref{eq:1.5}, in which ${y}^i_{a_0}$ and ${y}^i_{a_1}$ denote the segmentation mask on $i^{th}$ frame from non-key action $a_0$ and key action $a_1$ respectively, and $Acc({y}^i_{a_0}, {y}^i_{a_1})$ stands for the accuracy score with ${y}^i_{a_0}$ as the prediction and ${y}^i_{a_1}$ as the label. 
\begin{align}
\begin{split}
\label{eq:1.5}
\displaystyle
r_i &= 
    \begin{cases}
        0, & a_j=a_0.\\
        1 - Acc({y}^i_{a_0}, {y}^i_{a_1}), & a_j=a_1. \\
    \end{cases} \\
\end{split}
\end{align}

\textbf{Constraining key selection frequency} \hspace{0.1cm} The constraints on key selection frequency are necessary in our task. Since a key action will generally lead to a better reward than a non-key one, the policy network inclines to make all-key decisions if no constraint is imposed on the frequency of key selection. In this paper, we propose a \emph{stop immediately exceeding the limitation} approach. Particularly, for one episode consisting of $M+1$ frames $\{\mathbf{I}_t,\mathbf{I}_{t+1},...,\mathbf{I}_{t+M}\}$, the agent starts from $\mathbf{I}_t$ and explores continuously towards $\mathbf{I}_{t+M}$. At each time step, if the KAR in decision history has already surpassed a limit $\eta$, the agent will stop immediately and thus this episode ends, otherwise, it will continue until reaching the last frame $\mathbf{I}_{t+M}$. By using this strategy, a policy network should limit the use of key decision to avoid an over-early stopping, and also learn to allocate the limited key budgets on the frames with higher rewards. By varying the KAR limit $\eta$, we could train $\pi_{\theta}$ with different key decision frequencies.

\textbf{Episode settings} \hspace{0.1cm} Real-time videos usually contains enormous number of high-dimensional frames, thus it is impractical to include all of them in one episode, due to the high computational complexity and possible huge variations across frames. For simplicity, we limit the length of one episode $\{\mathbf{I}_t,\mathbf{I}_{t+1},...,\mathbf{I}_{t+M}\}$ to 270 frames (9 seconds) for 300VW and 30 frames (snippet length) for Cityscapes respectively. We vary the starting frame $\mathbf{I}_t$ during training to learn the global policies across videos. For each episode, we let the agent run $K$ times (with the aforementioned key constraint strategy) to obtain $K$ trials to reduce variances. The return of each episode can be expressed as $J(\theta)=\frac{1}{K}\sum_{v=1}^{K}\sum_{u=t}^{t+p_v}\gamma^{u-t}r_u^v$, where $t$ is the starting frame index of the episode,  and $p_v$ denotes the total step number at the $v^{th}$ trail (since agent may stop before $M$ steps), and $r_u^v$ refers to the reward of frame $u$ in $v^{th}$ trail. $J(\theta)$ is the main objective function to optimise.

\textbf{Auxiliary loss} \hspace{0.1cm} In addition to optimise the cumulative reward $J(\theta)$,  we employ the entropy loss $\mathcal{H}(\pi_{\theta}(\mathbf{a}|\mathbf{s}))$ as in \cite{mnih2016asynchronous,pang2018meta} to promote the policy that retains high-entropy action posteriors so as to avoid over-confident actions. Eq. \ref{eq:2} shows the final objective function $\mathcal{L}$ to optimise using policy gradient with reinforcement method \cite{williams1992simple}.

\begin{align}
\begin{split}
\label{eq:2}
\displaystyle
\mathcal{L} = J(\theta) + \lambda_1 \mathcal{H}(\pi_{\theta}(\mathbf{a}|\mathbf{s}))
\end{split}
\end{align}

\textbf{Epsilon-greedy strategy} \hspace{0.1cm} During training, agent may still fall into over-deterministic dilemmas with action posteriors approaching nearly 1, even though the auxiliary entropy loss have been added. To recover from such dilemma, we implement a simple strategy similar to epsilon-greedy algorithm for action sampling, i.e., in the cases that action probabilities $\pi_{\theta}(a_j|\mathbf{s})$ exceed a threshold $\epsilon$ (such as 0.98), instead of taking action ${a_j}$ with probability $\pi_{\theta}({a_j}|\mathbf{s})$, we use $\epsilon$ to stochastically pick action ${a_j}$ (and $1.0-\epsilon$ for picking action ${a_{1-j}}$).

\section{Experiments}
\label{Experiments}

\subsection{Datasets}
We conducted experiments on two datasets: the 300 Videos on the Wild (300VW) dataset \cite{shen2015first} and the Cityscapes dataset \cite{cordts2016cityscapes}. 300VW is used for evaluating the proposed real-time face segmentation system with the RL key selector. To the best of our knowledge, 300VW is the only publicly available face video dataset that provides per-frame segmentation labels. Therefore, to demonstrate the generality of our method, we also evaluate our method on Cityscapes \cite{cordts2016cityscapes}, which is a widely used dataset for scene parsing, and thus we show how our RL key scheduler can generalise to other datasets and scenarios.  

The 300VW dataset contains 114 face videos (captured at 30 FPS) with an average length of 64 seconds, all of which are taken in unconstrained environment. Following \cite{wang2018face}, we have cropped faces out of the video frames and generated the segmentation labels with facial skin, eyes, outer mouth and inner mouth for all the 218,595 frames. For experiment purpose, we divided the videos into three subject-independent parts, namely $A/B/C$ sets containing 51 / 51 / 12 videos. In detail, for training $\mathcal{N}$, we randomly picked 9,990 / 1,0320 / 2,400 frames from sets $A/B/C$ for training/validation/testing. To train $\mathcal{F}$, we randomly generate 32,410 / 4,836 / 6,671 \emph{key-current} image pairs with a varying gap between 1 to 30 frames from sets $A/B/C$ for training/validation/testing. We intentionally excluded set $A$ for policy network learning, since this set has already been used to train $\mathcal{N}$ and $\mathcal{F}$, instead, we used the full $B$ set (51 videos with 98,947 frames) for training and validating the RL key scheduler, and evaluated it on the full $C$ set (12 videos with 22,580 frames). 


The Cityscapes dataset contains 2,975 / 500 / 1,525 annotated urban scene images as training/validation/testing set, while each annotated image is the $20^{th}$ frame of a 30-frame (1.8 seconds) video snippet. To ensure a fair comparison on this dataset, we have adopted the same preliminary models ($\mathcal{N}$ and $\mathcal{F}$) and the model weights provided by the authors of DVSNet \cite{xu2018dynamic}, such that we only re-trained the proposed RL key schedulers using the Cityscapes training snippets. Following DVSNet \cite{xu2018dynamic}, our method and the baselines are evaluated on the validation snippets, where the initial frame is set as key and performances are measured on the $20^{th}$ annotated frame. 

\subsection{Experimental Setup}

\textbf{Evaluation metric} We employed the commonly used mean Intersection-over-Union (mIoU) as the evaluation metric. For the performance evaluation of different key schedulers, we measure: 1. the relationship between Average Key Intervals (AKI) and mIoU, as to demonstrate the effectiveness of key selections under different speed requirements, and 2. the relationship between the actual FPS and mIoU.  

\textbf{Training preliminary networks} \hspace{0.1cm}
On 300VW, we utilised the state-of-the-art Deeplab-V3+ architecture \cite{deeplabv3plus2018} for image segmentation model $\mathcal{N}$, and we adopted the FlowNet2-s architecture \cite{ilg2017flownet} as the implementation of flow estimation function $\mathcal{F}$. For training $\mathcal{N}$, we initialised the weights using the pre-trained model provided in \cite{deeplabv3plus2018} and then fine-tuned it. We set the output stride and decoder output stride to 16 and 4, respectively. We divided $\mathcal{N}$ into $\mathcal{N}_{feat}$ and $\mathcal{N}_{task}$, where the output of $\mathcal{N}_{feat}$ is the posterior for each image pixel, we then fine-tuned the FlowNet2-s model $\mathcal{F}$ as suggested in \cite{zhu2017deep} by freezing $\mathcal{N}_{feat}$ and $\mathcal{N}_{task}$. Also, we used the pre-trained weights provided in \cite{ilg2017flownet} as the starting point of training $\mathcal{F}$. The input sizes for $\mathcal{N}$ and $\mathcal{F}$ are both set to 513*513.

On Cityscapes, we have adopted identical $\mathcal{N}$ and $\mathcal{F}$ architectures as DVSNet \cite{xu2018dynamic} and directly use the weights provided by the authors, such that we only re-trained the proposed policy key scheduler. Besides, we have also adopted the frame division strategy from DVSNet and have divided the frame into four individual regions. We refer interested readers to \cite{xu2018dynamic} for more details.

\textbf{Reinforcement learning settings} \hspace{0.1cm} For state $\mathbf{s}_i=\{\mathcal{D}_{i \rightarrow k},\mathcal{E}_{i}\}$, following DVSNet \cite{xu2018dynamic}, we leveraged the features from the Conv6 layer of the FlowNet2-s model as the deviation information $\mathcal{D}_{i \rightarrow k}$, and we obtained the expert information $\mathcal{E}_{i}=\{KAR, LKD\}$ from the last 90 decisions. During the training of policy network,  $\mathcal{N}_{feat}$, $\mathcal{N}_{task}$ and $\mathcal{F}$ were frozen to avoid unnecessary computations. We chose RMSProp \cite{tieleman2012lecture} as the optimiser and set the initial learning rate to 0.001. The parameters $\lambda_1$ in Eq. \ref{eq:2} were set to 0.14. We empirically decided the discount factor $\gamma$ to be 1.0, as the per frame performance was equally important in our task. The value of epsilon $\epsilon$ in epsilon-greedy strategy was set to 0.98. During training, we set the threshold value $\tau$ for determining the key action to 0.5. We used the reward formulation as defined in Eq. \ref{eq:1} for 300VW. For Cityscapes, the modified reward as defined in Eq. \ref{eq:1.5} was used because most frames in the Cityscapes dataset are not annotated. The maximum length of each episode was set to 270 frames (9 seconds) for 300VW and 30 frames (snippet length) for Cityscapes respectively, and we repeated a relatively large number of 32 trials for each episode with a mini-batch size of 8 episodes for back-propagation in $\pi_{\theta}$. We trained each model for 2,400 episodes and validated the performances of checkpoints on the same set. We also varied the KAR limit $\eta$ to obtain policy networks with different key decision tendencies. 

\textbf{Baseline comparison} \hspace{0.1cm}  We compared our method with three baseline approaches on both datasets: (1) The adaptive key decision model DVSNet \cite{xu2018dynamic}; (2) The adaptive key scheduler using flow magnitude difference in \cite{xu2018dynamic}; (3) Deep Feature Flow (DFF) with a fixed key scheduler as in \cite{zhu2017deep}. 
We utilised the same implementations and settings for the baselines as described in DVSNet paper, and we refer the readers to \cite{xu2018dynamic} for details. Note that for the implementation of DVSNet on Cityscapes, we directly used the model weights provided by the authors, but we have re-trained the DVSNet model on 300VW. For our method, to obtain key decisions with different Average Key Intervals, we have trained multiple models with various KAR limit $\eta$, and also varied the key threshold values $\tau$ of those models.

\textbf{Implementation} \hspace{0.1cm} We implemented our method in Tensorflow \cite{tensorflow2015-whitepaper} framework. Experiments were run on a cluster with eight NVidia 1080 Ti GPUs, and it took approximately 2.5 days to train a RL model per GPU.

\begin{table}[t!]
\centering
\caption{The performances of various image segmentation models and the flow estimation model FlowNet2-s. For training FlowNet2-s, Deeplab-V3+ with ResNet-50 backbone is used as the key feature extractor $\mathcal{N}_{feat}$. FPS is evaluated on a Nvidia 1080Ti GPU. `N/A' refers to ``Not Applicable''.}
\label{tab:1} 
\begin{tabularx}{\columnwidth}{ b s j k}
\hline
Model & Eval Scales & mIoU(\%) & FPS \\ \hline
FCN (VGG16)  & N/A & 63.54 & 45.5 \\ \hline
Deeplab-V2 (VGG16) &  N/A &  65.80 & 3.44 \\ 
\hline
\multirow{2}{*}{\shortstack{Deeplab-V3+ \\ (Xception-65)}} &  1.0  & 68.25  & 24.4    \\ \cline{2-4}
& 1.25, 1.75  & 68.98 & 6.4 \\ \hline

\multirow{2}{*}{\shortstack{
Deeplab-V3+ \\ (MobileNet-V2)}} & 1.0  & 67.07 & 58.8    \\ \cline{2-4}
&  1.25, 1.75  & 68.20 & 21.7 \\ \hline

\multirow{2}{*}{\shortstack{
Deeplab-V3+ \\ (ResNet-50)}} & 1.0  & 67.50 & 33.3    \\ \cline{2-4}
&  1.25, 1.75  & \textbf{69.61} & 10.1 \\ \hline
\hline 
FlowNet2-s & N/A & 64.13 & 153.8 \\ \hline
\end{tabularx}
\end{table}

\subsection{Results}
\textbf{Preliminary networks on 300VW} \hspace{0.1cm}
We evaluated five image segmentation models on 300VW dataset: FCN \cite{long2015fully} with VGG16 \cite{simonyan2014very} architecture, Deeplab-V2 \cite{chen2018deeplab} of VGG16 version, the Deeplab-V3+ \cite{deeplabv3plus2018} with Xception-65 \cite{chollet2017xception} / MobileNet-V2 \cite{mobilenetv22018} / RestNet-50 \cite{he2016deep} backbones. We have also tested two different eval scales (refer \cite{deeplabv3plus2018} for details) for Deeplab-V3+ model.  As can be seen from Table \ref{tab:1}, Deeplab-V3+ with ResNet-50 backbone and multiple eval scales (1.25 and 1.75) has achieved the best mIoU with an acceptable FPS, therefore we selected it for our segmentation model $\mathcal{N}$. Its feature extraction part $\mathcal{N}_{feat}$ was used to extract key frame feature in \emph{key-current} images pairs during the training of FlowNet2-s \cite{ilg2017flownet} model $\mathcal{F}$, whose performance was evaluated by the interpolation results on current frames. From Table \ref{tab:1} we can discover that the interpolation speed with $\mathcal{F}$ is generally much faster than those segmentation models at the cost of a slight drop in mIoU (from 69.61\% to 64.13\%). Under live video scenario, the loss of accuracy can be effectively remedied by a good key scheduler. 

\begin{figure*}[ht!]
  \begin{center}
  \includegraphics[width=.9\linewidth]{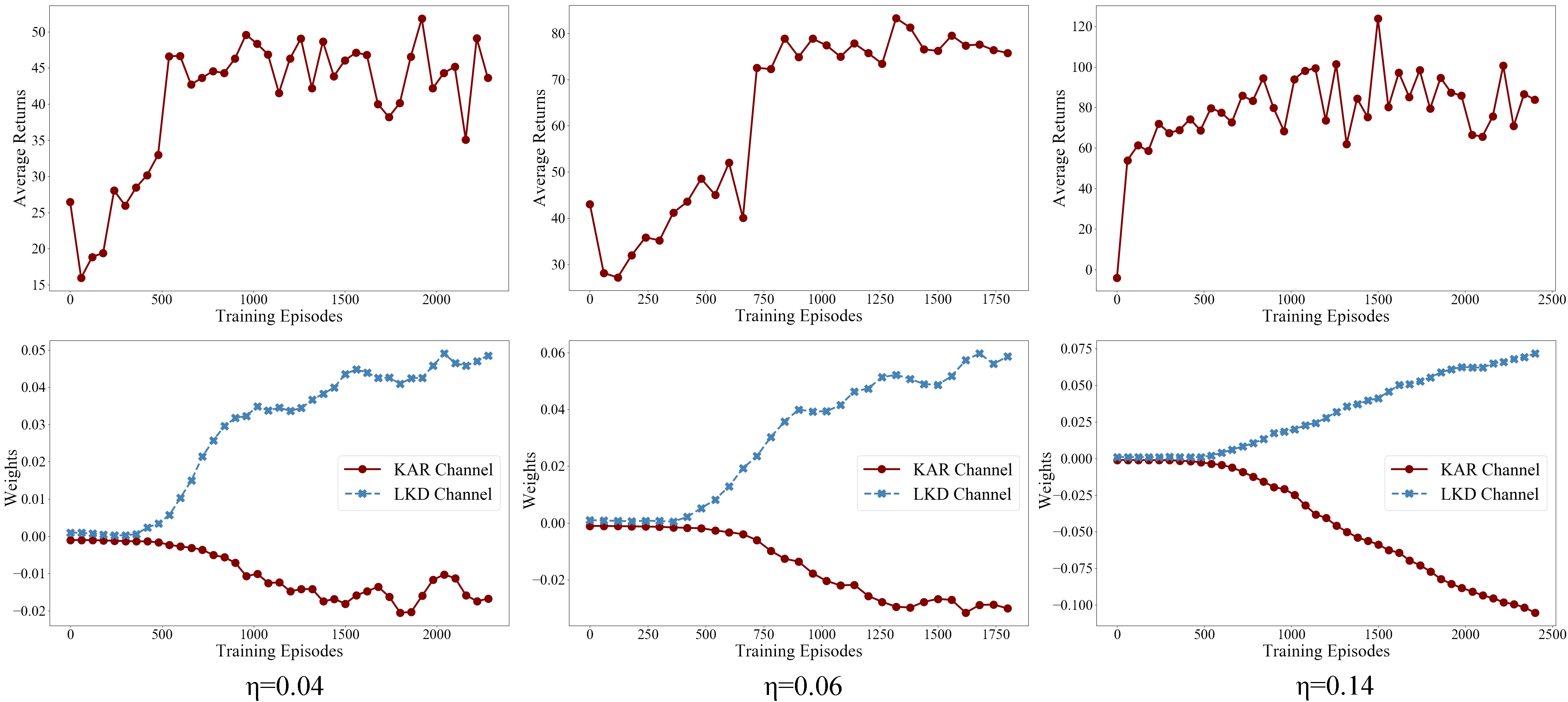}
  \end{center}
  \vspace{-2mm}
  \caption{The upper row plots the average return curves during RL training on 300VW with $\eta$ value set to 0.04, 0.06 and 0.14. The bottom row illustrates the variations of the weights of KAR and LDK channels contributing to the key posteriors $\pi_\theta(a_1|\mathbf{s})$ on the same dataset. The plots in the same column are from the same training session.}
  \label{fig:training_curve}
  \vspace{-2mm}
\end{figure*}

\textbf{RL training visualisation on 300VW} \hspace{0.1cm} In the upper row of Fig. \ref{fig:training_curve}, we demonstrate the average return during RL training with different KAR limits $\eta$ (0.04, 0.06, 0.14) on 300VW dataset. It can be seen that even though we select the starting frames of each episode randomly, those return curves still exhibit a generally increasing trend despite several fluctuations. This validates the effectiveness of our solutions for reducing variances and stabilising gradients, and it also verifies that the policy $\pi_\theta$ is improving towards more rewarding key actions. Besides, as the value of $\eta$ increases and allows for more key actions, the maximum return that each curve achieves also becomes intuitively higher.

We also visualised the influences of two expert information KAR and LDK by plotting their weights in $\pi_\theta$ during RL training on 300VW. In the bottom row of Fig. \ref{fig:training_curve}, we have plotted the weights of the two channels in $\pi_\theta$ that received KAR and LDK as input and contributed to the key posteriors $\pi_\theta(a_1|\mathbf{s})$, and we can observe that the weights of the LDK channel show a globally rising trend, while that of the KAR channel decrease continuously. Such trends indicate that the KAR/LDK channels become increasingly important in key decisions as training proceeds, since a large LDK value (or a small KAR) will encourage $\pi_\theta$ to take key action. This observation is consistent with the proposed key constraint strategy. Furthermore, we can also imply that the key scheduler relies more on  the LDK channel than the KAR with a lower $\eta$ like 0.04, conversely, KAR becomes more significant with a higher $\eta$ like 0.14.

\begin{figure}[t!]
  \begin{center}
  \includegraphics[width=0.72\linewidth]{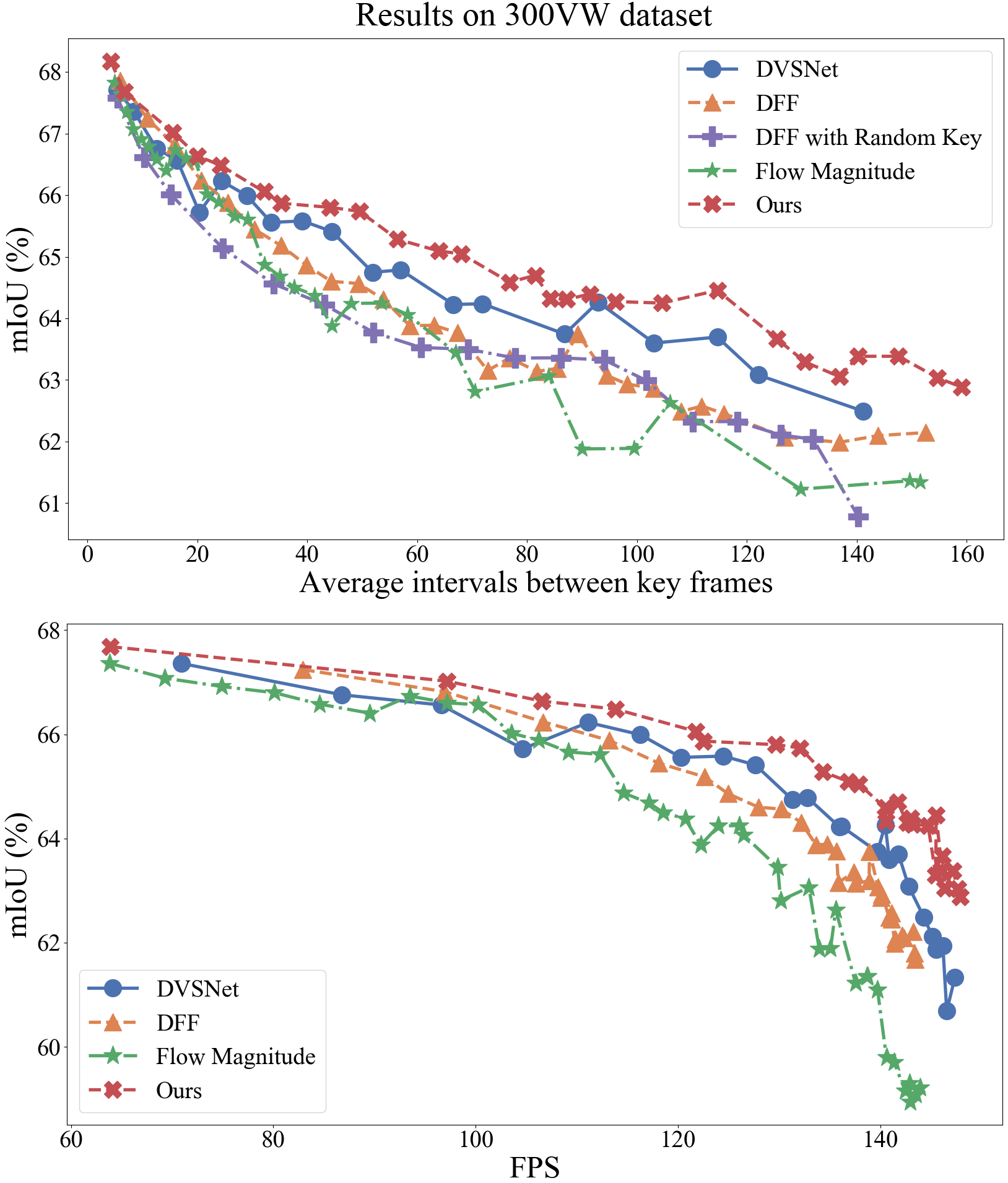}
   \end{center}
   \vspace{-2mm}
  \caption{Comparison between baselines and our approach on 300VW. Up: AKI versus mIoU, bottom: FPS versus mIoU. FPS is evaluated on a Nvidia 2080Ti GPU.}
  \label{fig:eval_results_300VW}
  \vspace{-2mm}
\end{figure}

\textbf{Performances evaluations} \hspace{0.1cm} The upper plot of Fig. \ref{fig:eval_results_300VW} shows the Average Key Intervals (AKI) versus mIoU of various key selectors on the 300VW dataset and the bottom plot depicts the corresponding FPS versus mIoU curves. Note that in the AKI vs. mIoU graph, we include two versions of DFF: the one with fixed key intervals and the variant with randomly selected keys. We can easily see that our key scheduler have shown superior performance than others in terms of both effective key selections and the actual running speed. Although the performance of all methods are similar for AKI less than 20, this is to be expected as the performance degradation on non-key frames can be compensated by dense key selections. Our method starts to show superior performance when the key interval increases beyond 25, where our mIoUs are consistently higher than that of other methods and decreases slower as the key interval increases.

\begin{figure}[t!]
  \begin{center}
  \includegraphics[width=0.72\linewidth]{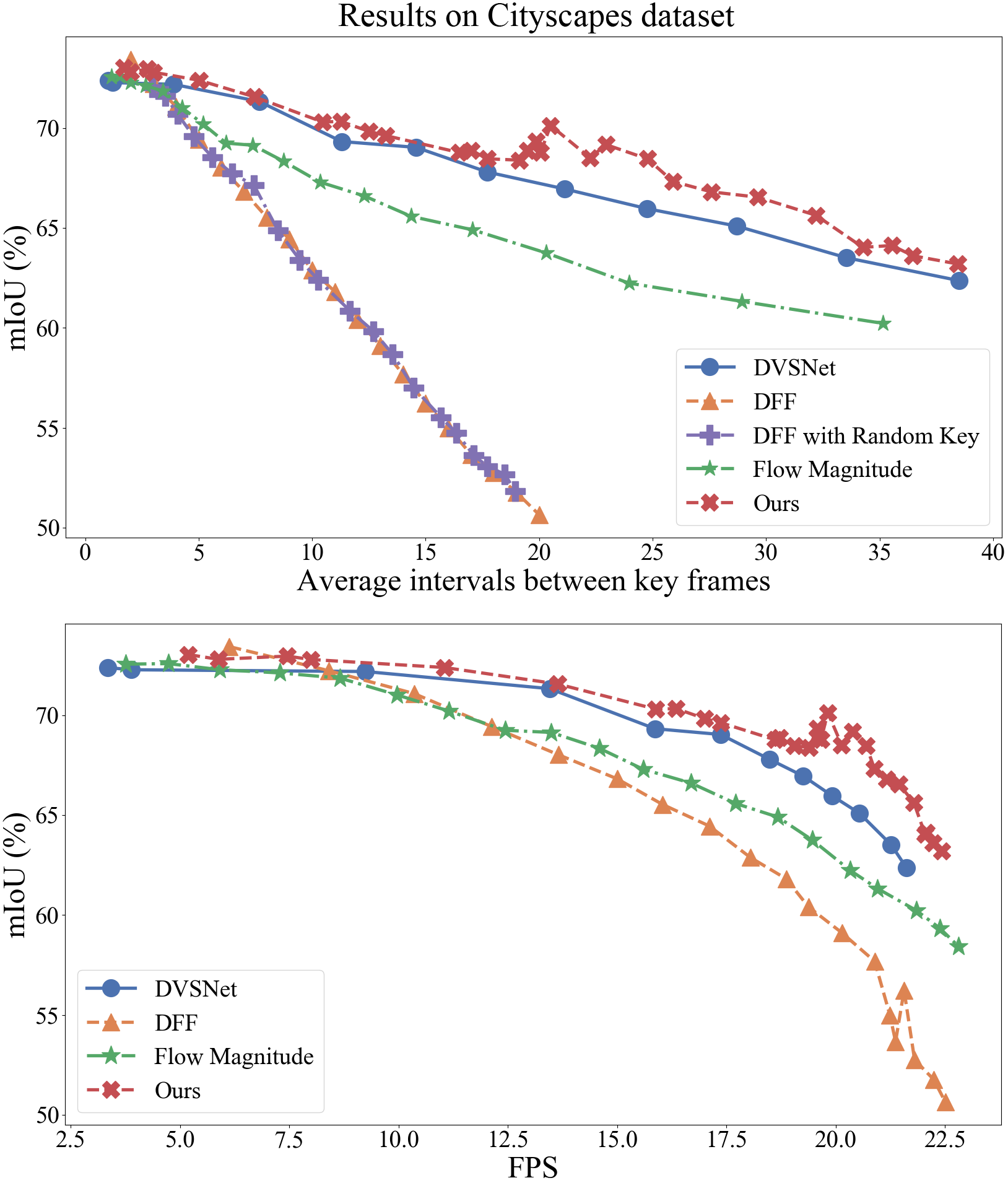}
   \end{center}
   \vspace{-2mm}
  \caption{Comparison between baselines and our approach on Cityscapes. Up: AKI versus mIoU, bottom: FPS versus mIoU. FPS is evaluated on a Nvidia 1080Ti GPU.}
  \label{fig:eval_results_cityscapes}
  \vspace{-2mm}
\end{figure}

The evaluation results on Cityscapes can be found in Fig. \ref{fig:eval_results_cityscapes}, which demonstrates a similar trend with those results on 300VW and therefore validates the generality of our RL key scheduler to other datasets and tasks. However, it should be noted that, in the case of face videos, selecting key frames by a small interval ($\ll20$) does not significantly affect the performance, which is not the same as in the autonomous driving scenarios of Cityscapes. This could be attributed to the fact that variations between consecutive frames in face videos are generally less than those in autonomous driving scenes. As a result, we can gain more efficiency benefit when using key scheduling policy with relatively large interval for dynamic segmentation of face video. 

\subsection{Visualising Key Selections}

\begin{figure}[ht!]
  \begin{center}
  \includegraphics[width=0.9\linewidth]{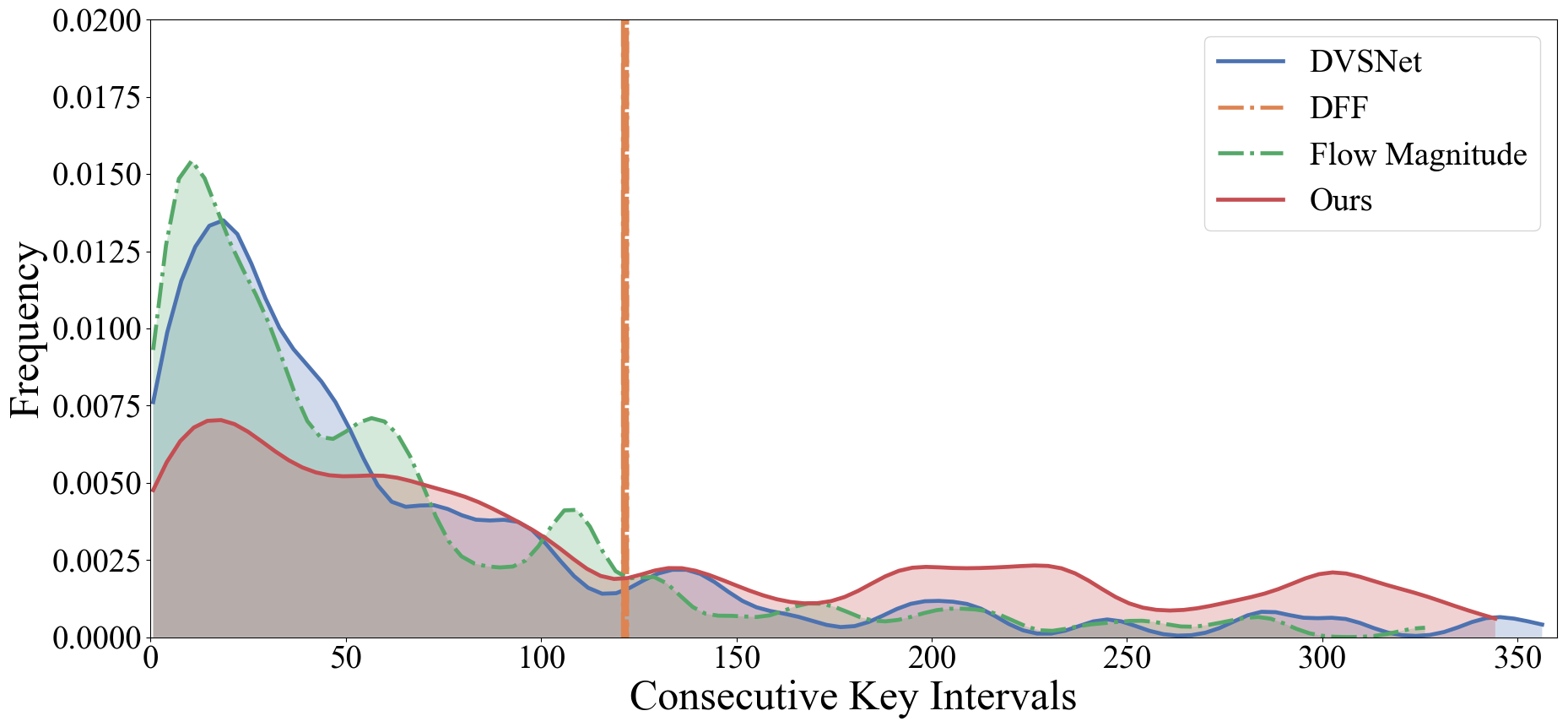}
  \end{center}
  \vspace{-2mm}
  \caption{The histogram plot for Consecutive Key Intervals of different methods on 300VW (AKI=121).}
  \label{fig:interval_histogram_plot}
  \vspace{-2mm}
\end{figure}

To better understand why our RL-based key selection method outperforms the baselines, we visualise the distribution of intervals between consecutive keys (CKI) based on the key selections made by all evaluated methods. Without loss of generality, Fig. \ref{fig:interval_histogram_plot} shows the density curves plotted from the experiment on 300VW dataset at AKI=121. As DFF uses a fixed key interval, its CKI distribution takes the shape of a single spike in the figure. In contrast, the CKI distribution given by our method has the flattest shape, meaning that the key frames selected by our method are more unevenly situated in the test videos. Noticeably, there are more cases of large gaps (>200) between neighbouring keys selected by our method than by others. This indicates our method could better capture the dynamics of the video and only select keys that have larger global impact to the segmentation accuracy.

\begin{figure}[t!]
  \begin{center}
  \includegraphics[width=0.92\linewidth]{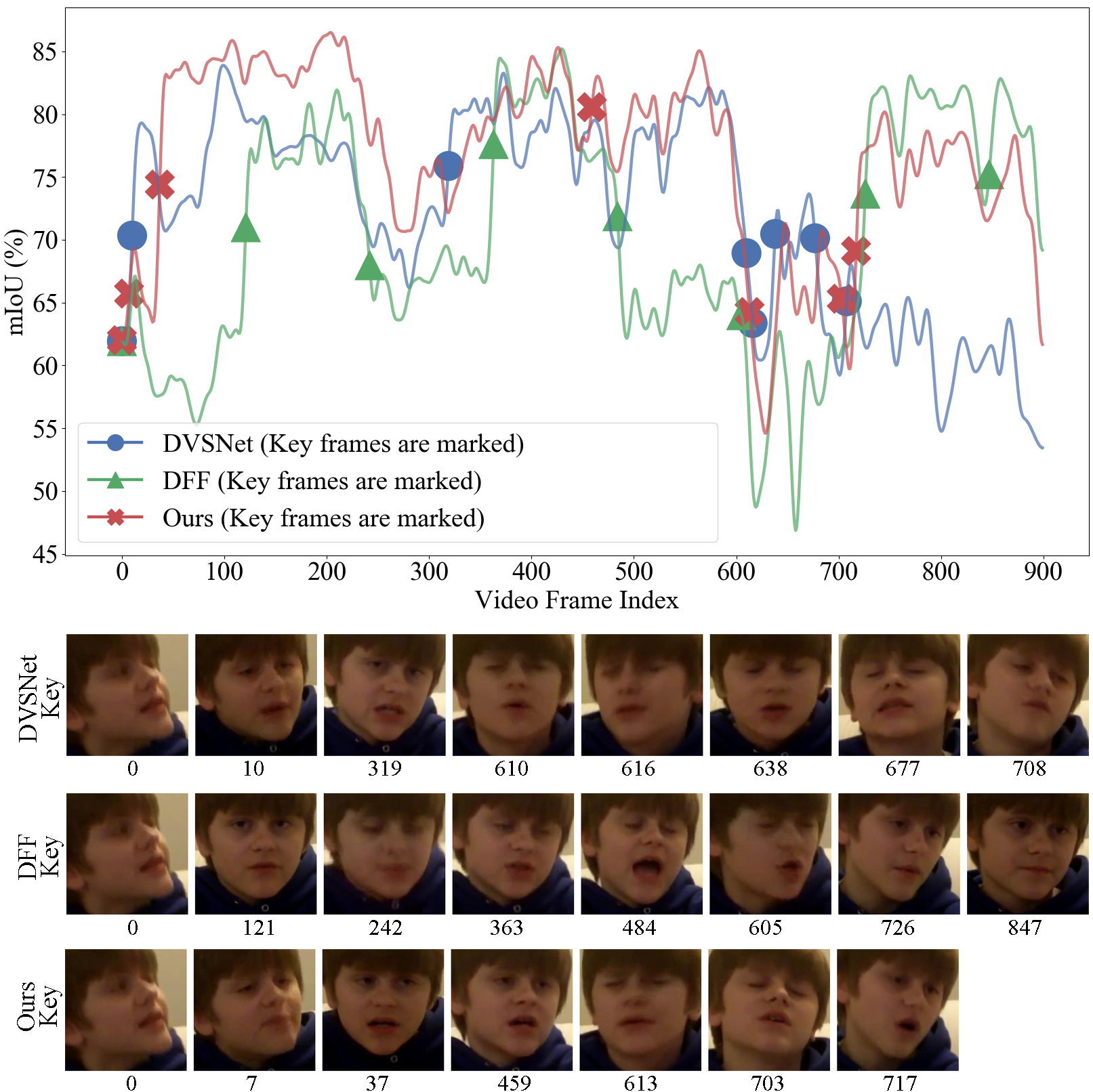}
  \end{center}
  \vspace{-2mm}
  \caption{A comparison of key selections on a 30-second face video between DVSNet, DFF and ours (AKI=121).}
  \label{fig:key_selection_plots}
  \vspace{-2mm}
\end{figure}

In addition, we also visualise the key frames selected by our method, DFF and DVSNet on a 30-second test video in Fig. \ref{fig:key_selection_plots} to provide insight on how the key selections can affect the mIoU. We can observe from this figure that the key frames selected by our method can better compensate for the loss of accuracy and retain higher mIoU over longer span of frames (such as frame 37 and 459), while those selected by DFF (fixed scheduler) are less flexible and the compensation to mIoU loss is generally worse than ours. Comparing DVSNet with ours, we can see that 1) our method can give key decisions with more stable non-key mIoUs (frames 37, 459 and 713), and 2) on hard frames such as frames 600 to 750, our reinforcement key scheduler has also made better compensations to performance loss with less key frames. These observations demonstrate the benefits brought by reinforcement learning, which is to learn key-decision policies from the global video context.  

\begin{figure}[t!]
  \begin{center}
  \includegraphics[width=0.9\linewidth]{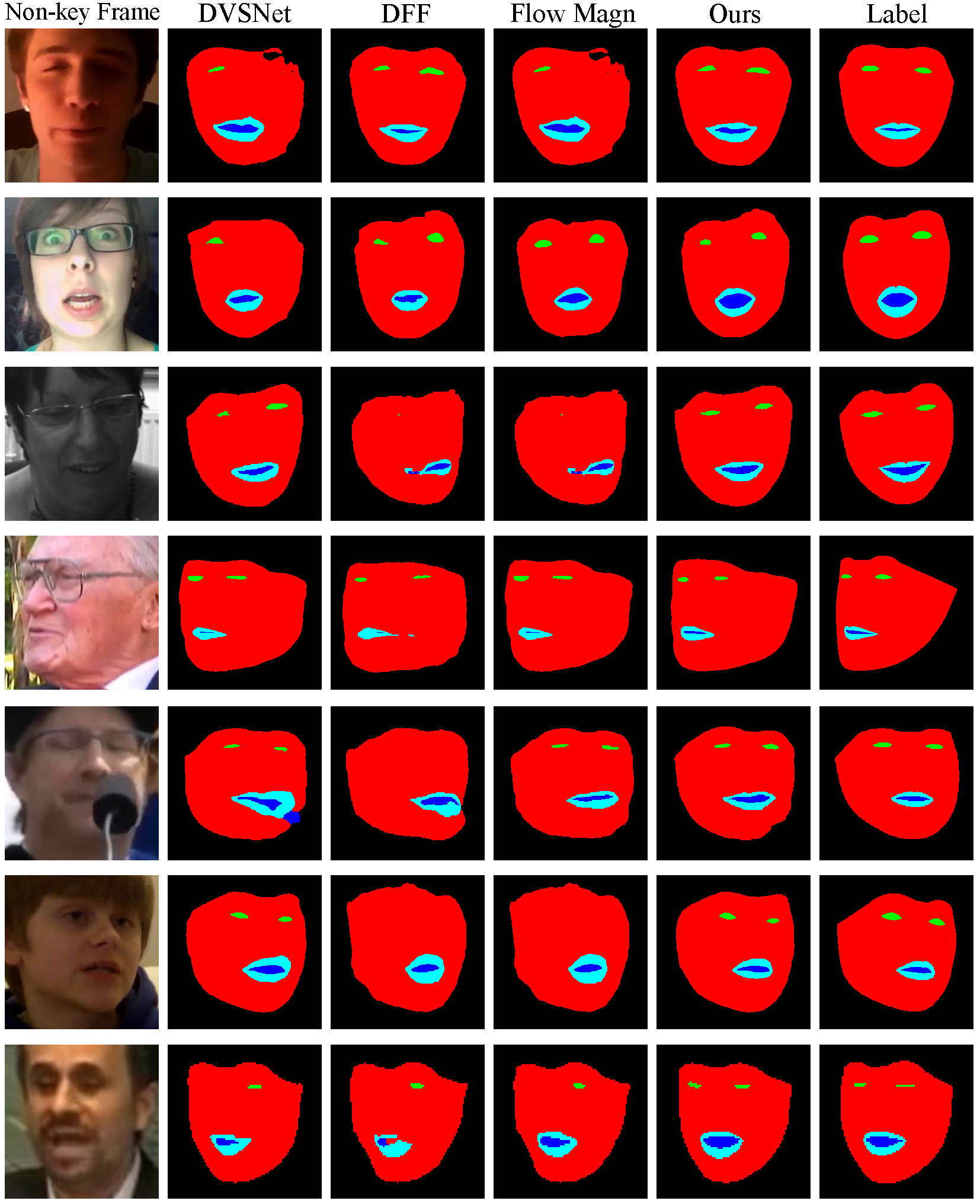}
  \end{center}
  \vspace{-2mm}
  \caption{The segmentation masks generated by different methods for the non-key frames on 300VW (AKI=121).}
  \label{fig:mask_visual_plot}
  \vspace{-2mm}
\end{figure}

Last but not least, in Fig. \ref{fig:mask_visual_plot}, we plot the segmentation masks generated by different methods on several non-key frames during the experiment on 300VW dataset (AKI=121). It can be seen that DFF with fixed key schedulers usually leads to low-quality masks with missing facial components, while the DVSNet and the Flow Magnitude methods have shown better but still not satisfying results. In contrast, our method has produced non-key masks with the best visual qualities, which further validate the effectiveness of the proposed key schedulers.

\section{Conclusions}
In this paper, we proposed to learn an efficient and effective key scheduler via reinforcement learning for dynamic face video segmentation. By utilising expert information and appropriately designed training strategies, our key scheduler achieves more effective key decisions than baseline methods at smaller computational cost. We also show the method is not limited to face video but could also generalise to other scenarios. By visualising the key selections made by our method, we try to explain why our key scheduler can make better selections than others. This is the first work to apply dynamic segmentation techniques with RL on real-time face videos, and it can be inspiring to future works on real-time face segmentation and on dynamic video segmentation. 

\section*{Acknowledgements}
The work of Yujiang Wang has been partially supported by China Scholarship Council (No. 201708060212) and the EPSRC project EP/N007743/1 (FACER2VM). The work of Yang Wu has been supported, in part, by Microsoft Research Asia through MSRA Collaborative Research 2019 Grant.

{\small
\bibliographystyle{ieee_fullname}
\bibliography{ref.bib}
}

\end{document}